\documentclass{article} 
\usepackage{nips12submit_e,times}
\usepackage{algorithm}
\usepackage{algorithmic}
\usepackage{psfrag,epsfig,subfigure}
\input{def.set}

\title{Learning Features with Structure-Adapting Multi-view Exponential Family Harmoniums}

\author{
Yoonseop~Kang$^1$ \qquad Seungjin Choi$^{1,2,3}$ \\
Department of Computer Science and Engineering$^1$, \\
Division of IT Convergence Engineering$^2$, \\
Department of Creative Excellence Engineering$^3$, \\
Pohang University of Science and Technology (POSTECH) \\
Pohang, South Korea, 790-784. \\
\texttt{\{e0en,seungjin\}@postech.ac.kr} \\
}

\nipsfinalcopy 

\begin{document}

\maketitle

\begin{abstract}
We propose a graphical model for multi-view feature extraction that automatically adapts its structure to achieve better representation of data distribution. The proposed model, {\em structure-adapting multi-view harmonium} (SA-MVH) has {\em switch parameters} that control the connection between hidden nodes and input views, and learn the switch parameter while training.
Numerical experiments on synthetic and a real-world dataset demonstrate the useful behavior of the SA-MVH, compared to existing multi-view feature extraction methods.
\end{abstract}

\section{Introduction}
Earlier multi-view feature extraction methods including canonical correlation analysis \cite{HardoonDR2004nc} and dual-wing harmonium (DWH) \cite{XingEP2005uai} assume that all views can be described using a single set of shared hidden nodes. However, these methods fail when real-world data with partially correlated views are given.
More recent methods like factorized orthogonal latent space \cite{SalzmannM2010aistats} or multi-view harmonium (MVH)  \cite{KangYS2011ecmlpkdd} assume that views are generated from two sets of hidden nodes: view-specific hidden nodes and shared ones. Still, these models rely on the pre-defined connection structure, and deciding the number of shared and view-specific hidden nodes requires a great human effort.

In this paper, we propose structure-adapting multi-view harmonium (SA-MVH) which avoids all of the problems mentioned above. Instead of explicitly defining view-specific and hidden nodes in prior to the training, we only use one set of hidden nodes and let each one of them to decide the existence of connection to views using {\em switch parameters} during the training. In this manner, SA-MVH automatically decides the number of view-specific latent variables and also captures partial correlation among views.

\section{The Proposed Model}
\label{sec:SA-MVH}

\begin{figure}
\center

\subfigure[DWH]{
\includegraphics[scale=0.5]{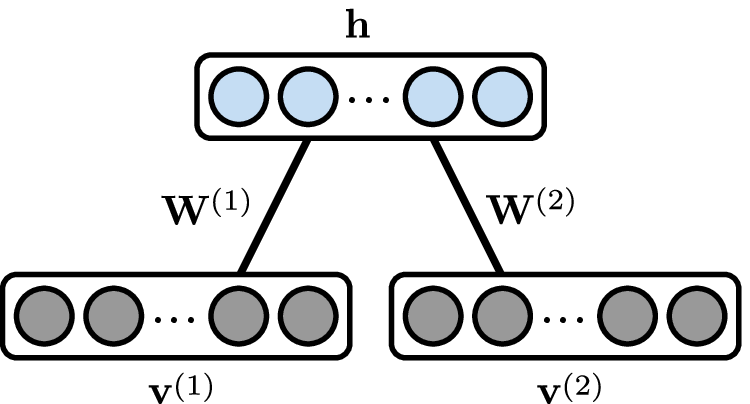}
} 
\subfigure[MVH]{
\includegraphics[scale=0.5]{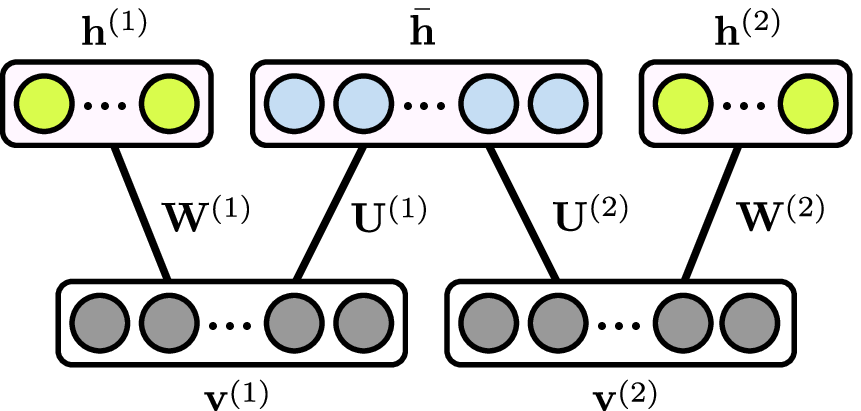}
}
\subfigure[SA-MVH]{
\includegraphics[scale=0.5]{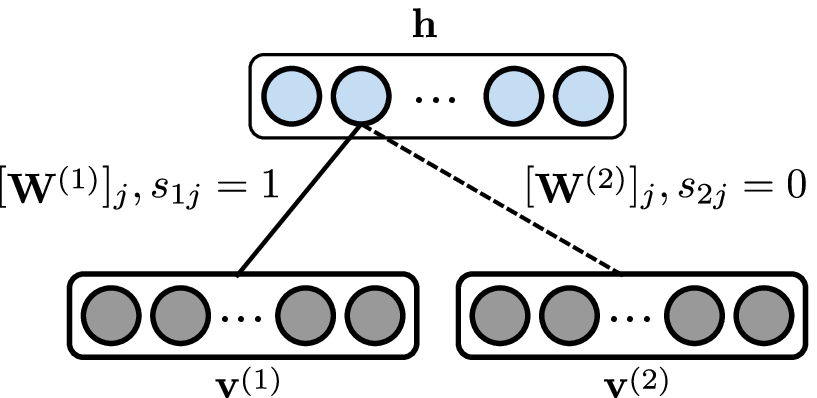}
}

\caption{Graphical models of (a) dual-wing harmonium, (b) multi-view harmonium, and (c) structure-adapting multi-view harmonium.}
\label{fig:gm}
\end{figure}

The definition of SA-MVH begins with choosing marginal distributions of visible node sets $\bv^{(k)}$ and a set of hidden nodes $\bh$ from exponential family distributions:
\begin{eqnarray}
p(v^{(k)}_i ) &\propto& \exp( \sum_a \xi^{(k)}_{ia} f^{(k)}_{ia}(v^{(k)}_i ) - A^{(k)}_i(\{ \xi^{(k)}_{ia} \})), \nonumber\\
p(h_j ) &\propto& \exp( \sum_b \lambda_{jb} g_{jb}(h_j) - B_j(\{ \lambda_{jb} \})), 
\end{eqnarray}
$f(\cdot)$, $g(\cdot)$ are sufficient statistics, $\xi$, $\lambda$ are natural parameters, and $A$, $B$ are log-partition functions.

Connections between visible nodes and hidden nodes of SA-MVH are defined by weight matrices $\{\bW^{(k)}\}$ and {\em switch parameters} $\sigma(s_{kj}) \in [0, 1]$, where $\sigma(\cdot)$ is a sigmoid function. A switch $s_{kj}$ controls the connection between $k$-th view and $j$-th hidden node by being multiplied to the $j$-th column of weight matrix $\bW^{(k)}$ (Figure \ref{fig:gm}). When $\sigma(s_{kj})$ is large ($ > 0.5$), we consider the view and the hidden node to be connected. With the quadratic term including weights and switch parameters, the joint distribution of SA-MVH is defined as below:
\begin{eqnarray}
p(\{\bv^{(k)}\}, \bh) 
\propto \exp\bigl( \sum_{k, i,j} \sigma(s_{kj}) \bW^{(k)}_{ij} f^{(k)}_i(\bv^{(k)}_i) g_{j}(h_j)
 - \sum_{k, i} \xi^{(k)}_i f^{(k)}_i(\bv^{(k)}_i)  -\sum_j \lambda_j g_j(h_j) \bigr) .
\end{eqnarray}
note that indices $a$ and $b$ are omitted to keep the notations uncluttered.

We learn the parameters $\bW^{(k)}$, $\xi^{(k)}$, $\lambda$, and switch parameters $s_{kj}$ by maximizing the likelihood of model via gradient ascent. 
The likelihood of SA-MVH is defined as the joint distribution of nodes summed over hidden nodes $\bh$:
\begin{eqnarray}
\mathcal L &=&  \langle \log p(\{\bv^{(k)}\}) \rangle_{data} =  \bigl\langle \log \sum_{\bh} p(\{\bv^{(k)}\}, \bh) \bigr\rangle_{data},
 \end{eqnarray}
 where $\langle\cdot\rangle_{data}$ represents expectation over data distribution.
 Then the gradient of log-likelihood with respect to the parameters $\bW^{(k)}$, $\xi^{(k)}$, $\lambda$, and $s_{kj}$ are derived as follows:
 \begin{eqnarray}
 {\partial \mathcal L \over \partial \bW^{(k)}_{ij}} &\propto&
 \bigl\langle \sigma(s_{kj}) f_i(\bv^{(k)}_i) B'_j({\hat \lambda_j}) \bigr\rangle_{data} 
 - \bigl\langle  \sigma(s_{kj}) f_i(\bv^{(k)}_i) B'_j({\hat \lambda_j}) \bigr\rangle_{model} \\
 {\partial \mathcal L \over \partial \xi^{(k)}_i} &\propto&
 \bigl\langle f^{(k)}_i(\bv^{(k)}_i)  \bigr\rangle_{data}
 - \bigl\langle f^{(k)}_i(\bv^{(k)}_i) \bigr\rangle_{model} \\
 {\partial \mathcal L \over \partial \lambda_j} &\propto&
 \bigl\langle B'_j({\hat \lambda_j})  \bigr\rangle_{data}
 - \bigl\langle B'_j({\hat \lambda_j})  \bigr\rangle_{model}, \\
 {\partial \mathcal L \over \partial s_{kj}} &\propto&
\left\langle \sigma'(s_{kj}) \bW^{(k)}_{ij} f_i(\bv^{(k)}_i) B'_j({\hat \lambda_j}) \right\rangle_{data}  - \left\langle \sigma'(s_{kj}) \bW^{(k)}_{ij} f_i(\bv^{(k)}_i) B'_j({\hat \lambda_j}) \right\rangle_{model}
 \end{eqnarray}
 where $\langle\cdot\rangle_{model}$ represents expectation over model distribution $p(\{\bv^{(k)}\}, \bh)$ and 
${\hat \xi^{(k)}_i} = \xi^{(k)}_i + \sum_j \sigma(s_{kj})\bW^{(k)}_{ij}g_j(h_j)$, 
${\hat \lambda_j} = \lambda_j + \sum_{k,i} \sigma(s_{kj})\bW^{(k)}_{ij}f_i(\bv^{(k)}_i)$ are shifted parameters.

\section{Numerical Experiments}
\label{sec:experiments}

\subsection{Feature Extraction on Noisy Arabic-Roman Digit Dataset}

\begin{figure}
\center
\subfigure[]{\includegraphics[width=0.18\textwidth]{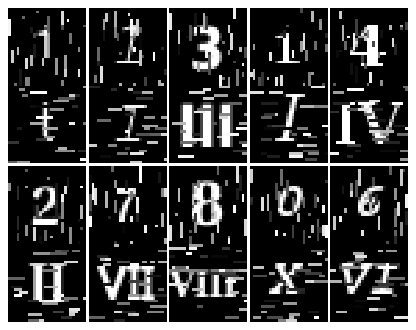} }
\subfigure[Shared features]{
\includegraphics[width=0.18\textwidth]{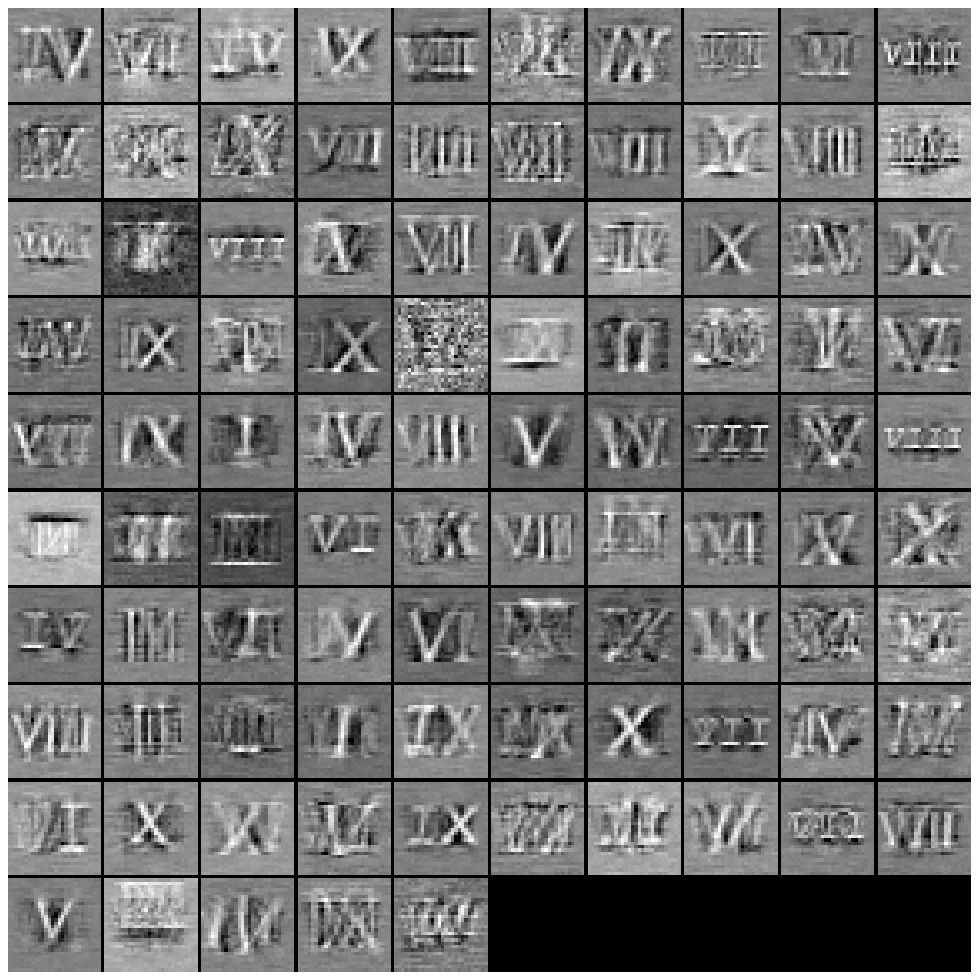}
\includegraphics[width=0.18\textwidth]{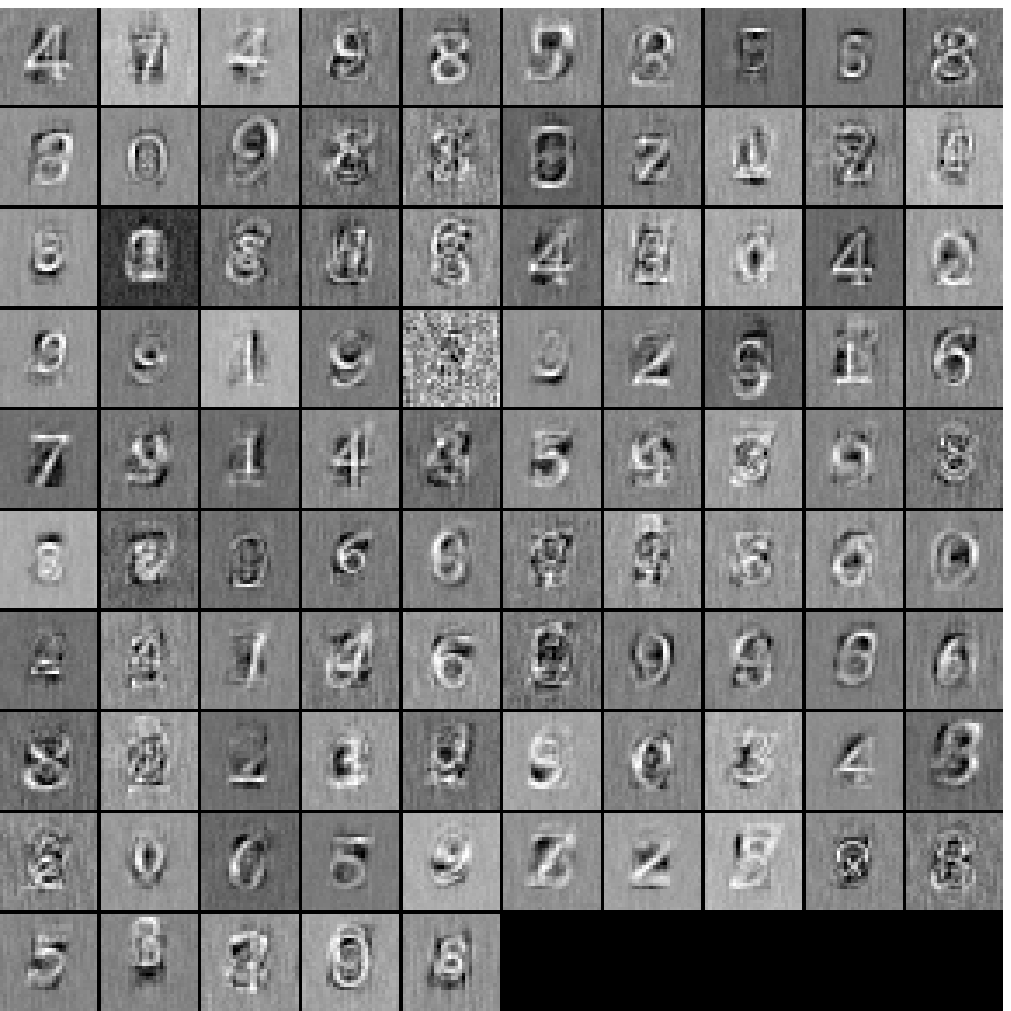}
}
\subfigure[View-specific features]{
\includegraphics[width=0.18\textwidth]{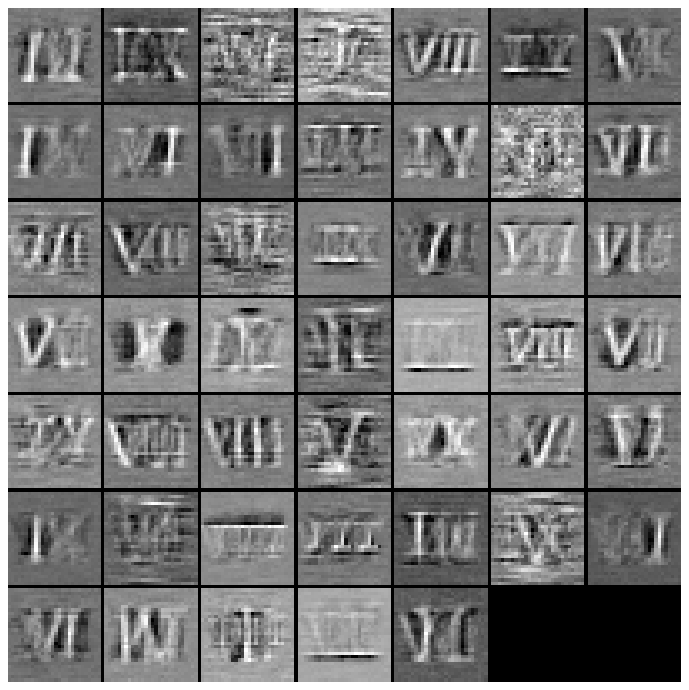}
\includegraphics[width=0.18\textwidth]{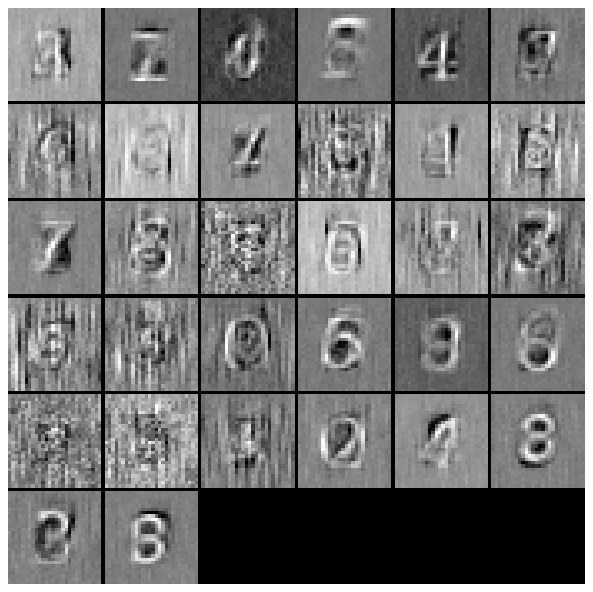}
}
\caption{(a) 10 samples from Noisy Arabic-Roman digit dataset, (b) shared features, and (c) view-specific features learned by SA-MVH.}
\label{fig:srbm_features}
\end{figure}

To simulate the view-specific and shared properties of multi-view data, we designed a synthetic dataset which contains 11,800 pairs of Arabic digits and the corresponding Roman digits written in various fonts. For each pair, we added random vertical line noises to Arabic digits, and horizontal line noises to Roman digits (Figure \ref{fig:srbm_features}-(a)).  SA-MVH trained with 200 hidden nodes found 95 shared features (with connection to both views), and 47 view-specific features for Roman digits, and 32 for Arabic digits. Remaining 26 were not connected to any views and ignored. Most of the shared features were noise-free and encoded parts of Roman and Arabic numbers (Figure \ref{fig:srbm_features}-(b)). On the other hand, the view-specific features had components with horizontal or vertical noises, as well as the  parts of the numbers (Figure \ref{fig:srbm_features}-(c)). In this example, SA-MVH automatically separated view-specific and shared information without any prior specification of the graph structure.

\subsection{Image Classification on Caltech-256 Dataset}
We extracted 512 dimensions of GIST features and 1,536 dimensions of histogram of gradients (HoG) features from Caltech-256 dataset to simulate multi-view settings. SA-MVH and other multi-view feature extraction methods based on harmonium -- DWH and MVH were trained on the dataset for comparison. 
We also compared our method to Sparse Filtering \cite{NgiamJ2011nips}, which is not a harmonium-based method.
We trained the feature extraction methods and tested the methods with k-nearest neighbor classifiers (Table \ref{table:result_caltech256}). 
SA-MVH resulted better than other feature extraction models in this experiment, regardless of the value of $k$ for nearest neighbor classifier. 

\begin{table}
\caption{Image classification accuracy of k-nn classifier using feature extraction methods trained on Caltech-256 dataset. For each value of $k$, the best result is marked as bold text.}\label{table:result_caltech256}
\center
\begin{tabular}{c||c|c|c|c|c}
\hline
Method & \# 10-NN & 30-NN & 50-NN & 70-NN & 100-NN \\
\hline
Sparse Filtering & 0.161 & 0.165 & 0.163 & 0.16 & 0.155 \\
DWH & 0.237 & 0.231 & 0.217 & 0.207 & 0.194 \\
MVH & 0.239 & 0.225 & 0.216 & 0.203 & 0.191 \\
SA-MVH & \textbf{0.246} & \textbf{0.232} & \textbf{0.223} & \textbf{0.212} & \textbf{0.198} \\
\hline
\end{tabular}
\end{table}

\section{Conclusion}
\label{sec:conclusion}

In this paper, we have proposed the multi-view feature extraction model 
that automatically decides relations between latent variables and input views.
The proposed method, SA-MVH models multi-view data distribution with less restrictive assumption 
and also reduces the number of parameters to tune by human hand.
SA-MVH introduces {\em switch parameters} that control the connections between hidden nodes and input views, and find the desirable configuration while training.
We have demonstrated the effectiveness of our approach by comparing our model to existing models in experiments on synthetic dataset, and image classification with simulated multi-view setting.

\bibliographystyle{IEEEtran}
\bibliography{sjc}

\end{document}